\documentclass{article}

\usepackage{arxiv}

\usepackage[utf8]{inputenc} 
\usepackage[T1]{fontenc}    
\usepackage{hyperref}       
\usepackage{url}            
\usepackage{booktabs}       
\usepackage{amsfonts}       
\usepackage{nicefrac}       
\usepackage{microtype}      
\usepackage{lipsum}
\usepackage{amssymb}
\usepackage{graphicx}
\usepackage{subcaption}

\title{Cost-sensitive Hierarchical Clustering for Dynamic Classifier Selection}

\author{
  Meinolf Sellmann \\
  GE Global Research\\
  \texttt{meinolf@gmail.com} \\
   \And
 Tapan Shah\\
  GE Global Research\\
  \texttt{tapan.shah@ge.com} \\
}
\begin{document}

\maketitle

\begin{abstract}
We consider the dynamic classifier selection (DCS) problem: Given an ensemble of classifiers, we are to choose which classifier to use depending on the particular input vector that we get to classify. The problem is a special case of the general algorithm selection problem where we have multiple different algorithms we can employ to process a given input. We investigate if a method developed for general algorithm selection named cost-sensitive hierarchical clustering (CSHC) is suited for DCS. We introduce some additions to the original CSHC method for the special case of choosing a classification algorithm and evaluate their impact on performance. We then compare with a number of state-of-the-art dynamic classifier selection methods. Our experimental results show that our modified CSHC algorithm compares favorably.  
\end{abstract}
\section{Introduction}

The idea of using more than one classifier to improve accuracy goes back to the basic theory of PAC learning~\cite{valiant84} and boosting weak learners~\cite{freund_short_nodate}. Often, we have multiple classifiers available to us, whereby these classifiers may be based on different concept classes or may themselves be ensembles. We could use a cross-validation to determine the best classifier and deploy it. We may be able to do even better, though, if we choose dynamically, after seeing the feature input, which classifier to use. That is, rather than choosing one classifier and using it regardless of the input, we may use one classifier for one input and another for another. A method is needed to choose a classifier. This problem is known in the literature as dynamic classifier selection (DCS). 

Of course, there are other ways to combine multiple classifiers, for example by using each classifier's support for each of the possible class labels and aggregating this information. This is the basic idea behind stacking~\cite{wolpert_stacked_1992}. Note that, in stacking, the final class label may not coincide with any of the classes chosen by any of the base classifiers in the ensemble, which gives this technique more flexibility and the potential to outperform dynamic classifier selection. However, the disadvantage of this more flexible aggregation method is that every classifier in the ensemble needs to score each possible class label. 
Another disadvantage of more elaborate aggregation schemes is that it makes explaining the classification more challenging. When using dynamic classifier selection, we can inherit the explanation method from the base classifier. In this paper, we therefore limit ourselves to dynamic classifier selection. 

\smallskip

A problem highly related to DCS has been identified in the satisfiability and optimization communities. It was found that different algorithmic approaches may solve the same problem instances in vastly different compute times. The idea arose to choose which algorithm to employ only after the concrete instance to process is known~\cite{leyton-brown_portfolio_2003}. These so-called "algorithm portfolios" have since led to a massive improvements in our ability to solve extremely hard combinatorial satisfiability and optimization problems~\cite{sat_competition}.

One method for selecting an algorithm out of a portfolio of algorithms was introduced in~\cite{malitskyAlg2013} and employs cost-sensitive multi-classification for algorithm selection. In this paper, we investigate whether this approach for general algorithm selection can be used effectively for DCS. We introduce several modifications to make the method more suited for classifier selection. Then, we compare the approach with state-of-the-art DCS methods. 

\section{Related Work on Dynamic Classifier Selection} 
State-of-the-art DCS methods work by estimating the competence of the base classifiers for a given query sample and then select the base classifier with highest competence. The competence is commonly estimated as follows:
\begin{enumerate}
    \item For a given sample, a region of competence i.e. a local neighborhood of training samples  is computed using either $k$-nearest neighbors ($k$-NN) or clustering methods.
    \item Then, the competence level of each classifier is computed on the neighborhood, based on varying criteria like accuracy of base classifiers, ranking, etc.
\end{enumerate}

Prominent examples that realize the framework above are Local Class Accuracy (LCA)~\cite{woods_combination_1997}, Overall Local Accuracy~(OLA)~\cite{woods_combination_1997}, A Priori~\cite{giacinto_methods_1999}, A Posterori~\cite{giacinto_methods_1999}, and Multiple Classifier Behavior~\cite{giacinto_dynamic_2001}: 
\newline
{\bf OLA}: In this approach,  the competence of a base classifier is defined as the overall accuracy of the classifier in the local neighborhood. The local neighborhood is extracted using $k$-NN, where $k$ is a tunable parameter. As all other approaches realizing this framework, the classifier with the highest competence in the local neighborhood is chosen. 
\newline
{\bf LCA}: This method is similar to OLA, with the difference that it uses a different notion of class-specific accuracy where only the accuracy of the class predicted by a classifier is considered. 
\newline
{\bf A Priori (APR) and A Posteriori (APO)}: Both these methods use the "soft" conditional class probabilities which are output by the base classifiers, instead of "hard" class predictions used in LCA and OLA to compute the probability of correct classification. Note that this implies that APR and APO are equally costly as stacking techniques. Both APR and APO use $k$-NN to define the local neighborhood. The key difference between APR and APO is that APR computes the competence of a base classifier without knowledge of the class predicted by the base classifier on the query sample. On the other hand, if the base classifier predicts class $C$ for a query sample, APO computes the competence by limiting to those samples in the local neighborhood with actual class $C$.\footnote{We found that APO performed consistently worse than APR and therefore do not include it in the results for APO in Table \ref{tab:results}.}
\newline
{\bf MCB}: A concept called Behavioral Knowledge Space (BKS) is used to refine the $k$-NN neighborhood: Only samples with similar "output profiles" are kept in the local neighborhood, whereby an output profile of an input feature vector is the vector of predictions of all the base classifiers. Class-specific accuracy is the used as competence score.

Methods that deviate slightly from the framework above select a subset of classifiers that perform well in terms of competence criteria on the neighborhood. A majority vote by the subset of classifiers is then conducted for the final prediction. Since the majority class must have at least one classifier that voted for it, we can select any such classifier, which is why these methods can also be viewed as dynamic classifier selection methods.

Examples of methods that realize this modified framework are $k$-Nearest Oracle Eliminate (KNORA-E), $k$-Nearest Oracle Union (KNORA-U)~\cite{ko_dynamic_2008}, and META-DES~\cite{cruz_meta_2015}.
\newline
{\bf KNORA-E (KE)}: Given a local neighborhood created using $k$-NN for a query sample, all base classifiers with less than 100\% accuracy on the neighborhood training samples are eliminated, whereby $k$ is reduced until this is the case for at least one classifier. As with all other methods realizing this framework, of the remaining classifiers, a majority vote is taken to arrive at the final prediction.
\newline
{\bf KNORA-U (KU)}: This method is similar to KNORA-E, with the difference that all classifiers which are correct for at least one training sample in the neighborhood are also retained. Moreover, voting is weighted: The weight of the vote by a base classifier is equal to the number of neighborhood samples that are classified correctly.\footnote{We found that KU significantly outperforms KE which is why we do not include results on KE in Table~\ref{tab:results}.}
\newline
{\bf META-DES (MD)}: This method trains a meta-classification algorithm on a set of meta-features, where the meta-classes are "competent" or "incompetent," to select the set of classifiers that vote by simple majority on the final class. The meta-features include class-specific and overall accuracy in the local neighborhood, classifier probability, classifier consensus, and some others. The local neighborhood is extracted using BKS. To train the meta-classifier, whenever a classifier labels a training input correctly, it is labeled competent, and incompetent otherwise.

In our experiments, we also compare with the simplest (static) method used in multiple classification systems, whereby all the base classifiers are pooled and the output is obtained using the majority voting (MV) rule.

For further details, we refer to the very thorough discussion of various dynamic classifier and ensemble selection methods in~\cite{cruz_dynamic_2018}.

\section{Cost-sensitive Hierarchical Clustering}

The main objective of this paper is to study the effectiveness of "cost-sensitive hierarchical clustering" (CSHC) for DCS. The idea behind CSHC is simple: Recursively split a cluster of input samples such that the inputs within a partition can agree on one algorithm that shall be used to process all inputs in the respective partition. 
In the original CSHC paper, the authors experimented with different ways how to split a cluster. In the end it was found effective and simple to consider recursively splitting clusters by selecting one feature and associated splitting value and to put all examples that have a respective feature value lower or equal the splitting value in one sub-cluster, and the others in the other. That is to say, the final version of CSHC essentially builds a decision tree. However, it does not use entropy to determine splitting features and values. Instead, CSHC considers the overall performance when using a different, optimal algorithm for each partition, rather than the same algorithm on all examples in the parent cluster. The split that results in the best performance gain is then selected. 

Note that performance can be any metric desired, from running time (which is typically the target in search and optimization), to optimality gap within a fixed time frame (a typical metric when tuning local search heuristics), to some other metric of quality. For the purpose of classifier selection, we will simply use the number of input samples that a classifier labels correctly, i.e., the method's accuracy. 

Three hyper-parameters guide when the recursive splitting of clusters stops. The first is a simple depth limit, the second a minimum number of samples that must remain in each cluster, and the last is a minimum improvement that is expected from splitting a cluster.

As it is the case with decision trees, it has been found beneficial to build more than one hierarchical clustering. Identically to how random forests work, in CSHC, for each new clustering, only a subset of features are allowed to be used to split the inputs, and a sub-sample (with replacement) is built from the total set of inputs to be clustered. 

Three hyper-parameters guide this process of ensembling clusterings: How many clusterings (trees) to construct, how many features are randomly selected to be used for splitting the sample set, and how often we sample the training inputs with replacement. 

This concludes the description of the training phase of CSHC. When using the clusterings to choose an algorithm for a new input, we require a process that resolves conflicts between the recommendations from different clusterings. Various methods have been described in the original CSHC paper~\cite{malitskyAlg2013}. Here, we will limit ourselves to the idea of using the algorithm that has the best cumulative rank over all clusterings. That is, when we are given a new input at test time, we determine which clusters the input falls into for each of the clusterings. Then, we rank all algorithms for each cluster. We select the algorithm that has the best cumulative rank when summing up the ranks over all clusters. For further details on CSHC, please see~\cite{malitskyAlg2013}.

\smallskip

Note how CSHC differs from existing DCS methods. Superficially, one might think that CSHC also builds a neighborhood and then selects the best performing classifier on that neighborhood. However, the way how that neighborhood is constructed and how performance is assessed is very different. First, the multiple hierarchical clusterings built by sub-sampling the training samples with repetition creates neighborhoods (the multi-set of examples in clusters the target feature vector is assigned to) that give different weights to different training examples by including samples as many times as they appear in target clusters.  

Second, the clusterings are constructed not by considering unsupervised metric regions in the feature-space, or regions where the original machine learning problem favors the same class, but by considering regions which are handled well by the same classifier. And finally, the performance is assessed by ranking classifiers on multiple clusters and picking the best, which is unlike how any other existing method determines the final selection. 

\section{Modifications for Dynamic Classifier Selection}

CSHC can be applied to any algorithm selection problem and is hence directly applicable to DCS as well. However, certain aspects make classifier selection a special case of general algorithm selection. In this section, we discuss these differences and propose some modifications to the vanilla CSHC methodology.

\subsection{Training Data}
The first particularity of DCS is the way how the training data is generated. When building an algorithm selector for an optimization problem, for example, we simply run the various algorithms on each training instance and thereby gather the cost data needed to train the clusterings with CSHC. That is to say that, in other applications, the training instances used to train the selector usually have no influence on the algorithms in the portfolio.  

When using an algorithm selector for classifier selection, this is not so clear anymore. There is a certain amount of labeled (with the classes of the original machine learning problem) data available, and this data needs to be used for training the base classifiers as well as the classifier selector (whereby the labels are used to determine the associated cost of each classifier). Obviously, the selector could be over-confident with a classifier if it only had access to cases where the classifier labels samples that were used to train the respective classifier. To circumvent this issue, we conduct a three-fold cross validation. In each fold, we use two thirds of the training data to train a classifier, then we evaluate the classifier on the remaining third of the data. The cost labels generated for CSHC are then exclusively derived from the validation performances. Note that, in this way, we can use the entire training data for the generation of clusterings.

\subsection{CSHC-Rank Regression}

Another difference is that, in general algorithm selection, we cannot always run all algorithms. Imagine a case where we need to choose the best scheduler for a given scheduling instance. 'Best' in this case usually means 'fastest.' Running all schedulers is obviously not an option, we have to choose one before we see the algorithm output. 

In the context of ensemble learning, the situation may be different. Of course, there may be scenarios where running all classifiers is too costly, for example because of latency requirements or because it is simply too cost-prohibitive to run them all. In this case, we can simply use the cumulative ranking procedure from CSHS. We will report on the performance of this method in the experimental results.

In other cases, however, our prime concern is classification performance rather than computational cost. Then, we may want to run all classifiers and use the classification results as well as the original features to select a classifier (and the associated class this classifier labels the input with). In the following, we present methods how to use this information in the context of CSHC.

One way how we can use the labels produced by the different classifiers is by voting. To this end, each classifier is assigned a certain weight, and the class it labels the input with gets this weight added as support. We select one of the classifiers that labels the input with the class that has the most support. Among all classifiers that lend the support, we select the one with the largest weight (with ties broken randomly).

The question is what weight to assign to each classifier. We utilize the ranks that CSHC provides for this purpose. Particularly, we assign the cumulative rank over all clusterings (with the better classifiers having higher rank) as the weight for each classifier. Note that the clusters considered are input--specific. Therefore, the weight each classifier is assigned changes dynamically from test sample to test sample.

\subsection{Linear Programming-based Weighting}

Rather than using cumulative ranks as weights, we can also employ a more labor-intensive method and compute a set of weights that would optimize the performance over the multi-set of samples over all clusters the given feature vector is assigned to by CSHC. We propose to set up a linear program (LP) for this purpose.

Assume we are given the number $n\in \mathbb{N}$ of different classifiers in the ensemble, the number $C\in \mathbb{N}$ of classes, the set $E=\{e_1,\dots,e_k\}$ of unique samples in the union of all clusters the given feature vector falls into, the correct labels $y_i\in\{1,\dots,C\}$, as well as numbers $m_i\in\mathbb{N}$ for $1\leq i\leq k$ that determine how often example $e_i\in E$ appears in the multi-set of samples returned by CSHC of the given feature vector. Finally, assume that, for each classifier $a\in\{1,\dots,n\}$ and each sample $e_i\in E$, we are given the label $l_i^a\in\{1,\dots,C\}$.

The LP we set up has three sets of variables. First, for each classifier $a\in\{1,\dots,n\}$, a weight $0\leq w_a\leq 100$ with $a\in\{1,\dots,n\}$. Moreover, for each unique example $e_i\in E$, we introduce two penalty variables $g_i,f_i\geq 0$. We impose the following constraints: First, the weight variables must sum to 100: $\sum_{c\leq C} w_c =100$. Next, for each unique example $e_i\in E$ and each class $c\in\{1,\dots,C\}$ with $c\neq y_i$, we add two constraints: $g_i + \sum_{a,l_i^a=y_i} w_a - \sum_{b,l^i_b=c} w_b \geq \gamma$ and $f_i + \sum_{a,l_i^a=y_i} w_a - \sum_{b,l^i_b=c} w_b \geq 1$. Then, we solve the LP to obtain weights and penalties that minimize the total penalty $\sum_{i\leq k} m_i (g_i + 2 f_i)$.

The LP aims to find a weighting for the classifiers such that the support for the correct label is at least $\gamma$\% more than the maximal support for any other label over the multi-set of examples that was returned by CSHC for the given feature vector. When that is not possible, the LP will strive to have at least the largest support for the correct label, or to get as close to the largest support as possible. For each example for which the weighted aggregate results in a class label that is correct, there is no penalty. Otherwise, the penalty is two times the gap of the total support for the wrong label minus the support for the correct label, plus whatever is needed to bring the gap between the support for the correct class to any other class to at least $\gamma$.

As previously, based on the weights obtained, we compute the class that has the most aggregate support and ultimately select the classifier that has the maximum weight among all that label the input with that maximally supported class.

\subsection{Confidence Assessment and Recourse}

We now have three different selection methods: The original CSHC which chooses the classifier with the highest cumulative rank over all clusters, rank-weighted voting, and finally by optimizing the aggregation weights via linear programming. We will investigate each one of these methods in the numerical results section. 

A more robust selection mechanism, inspired by \cite{AnsoteguiST18},  may be obtained by employing a process that considers how confident each of these classier selection methods actually is. To this end, for the two methods introduced above, we consider the ratio between the class that has the second largest support and the class that has the largest support (based on the respective ways to compute the weights of each classifier, either by cumulative rank or by solving the linear program). The lower this ratio (we refer to this parameter with $\rho$), the higher our confidence that this selection is correct. We propose to use rank regression first, since it is computationally much cheaper than solving a linear program for each test sample. If the confidence in the rank selection is high, we return the classifier selected. If the confidence does not exceed the given threshold, we next compute the classifier selected by the LP weighting scheme. We assess the confidence in this method as well. If confidence is high enough, we return the respective classifier. 

If confidence is also low in the LP-based weighting method, then we proceed as follows: If the class label of the classifier chosen by both rank regression and LP-based weighting are the same, then we return the classifier whose respective selection method has higher confidence (note that the class labels may be the same even when the two methods choose a different classifier). If the classes are not the same, we next check if the classifier returned by the original CSHC labels the input with the same class as one of the other two classifiers. If so, we return that classifier. If this also fails, which implies that all three selection methods select a different classifier and all three classifiers label the input with a different class, then we finally compute the dominant class label in the multi-set of training samples returned by CSHC. If one of the three classifiers provided by the three methods labels the input with that dominant class, then we choose this classifier. Otherwise, we return the classifier chosen by the LP-based weighting scheme.

\section{Numerical Results}
In this section, we describe the experiments to quantify the performance of our methods as well as  compare it against competing methods.
\subsection{Experiment Setup}
\subsubsection{Data Sets}
We use 40 data sets from OpenML~\cite{feurer_openml-python_2019,vanschoren_openml_2014} for our experiments. 
The details of the OpenML datasets used for numerical experiments are given in Table \ref{table:datasets}.

\begin{table}[b!]
    \centering
   \begin{tabular}{|l|c|c|c|c|}
\hline
           \bf name &  \bf \# features & \bf  \# classes &  \bf \bf \# training samples &  \bf \# test samples \\ \hline
         heart-h &          13 &          2 &                 196 &              98 \\ \hline
        credit-g &          20 &          2 &                 670 &             330 \\ \hline
     tic-tac-toe &           9 &          2 &                 641 &             317 \\ \hline
        kr-vs-kp &          36 &          2 &                2141 &            1055 \\ \hline
     qsar-biodeg &          41 &          2 &                 706 &             349 \\ \hline
            wdbc &          30 &          2 &                 381 &             188 \\ \hline
         phoneme &           5 &          2 &                3620 &            1784 \\ \hline
        diabetes &           8 &          2 &                 514 &             254 \\ \hline
 ozone-level-8hr &          72 &          2 &                1697 &             837 \\ \hline
     hill-valley &         100 &          2 &                 812 &             400 \\ \hline
             kc2 &          21 &          2 &                 349 &             173 \\ \hline
   eeg-eye-state &          14 &          2 &               10036 &            4944 \\ \hline
        spambase &          57 &          2 &                3082 &            1519 \\ \hline
             kc1 &          21 &          2 &                1413 &             696 \\ \hline
            ilpd &          10 &          2 &                 390 &             193 \\ \hline
             pc1 &          21 &          2 &                 743 &             366 \\ \hline
             pc3 &          37 &          2 &                1047 &             516 \\ \hline
        mozilla4 &           5 &          2 &               10415 &            5130 \\ \hline
           scene &         299 &          2 &                1612 &             795 \\ \hline
            musk &         167 &          2 &                4420 &            2178 \\ \hline
          letter &          16 &         26 &               13400 &            6600 \\ \hline
           nomao &         118 &          2 &               23091 &           11374 \\ \hline
   gina\_agnostic &         970 &          2 &                2323 &            1145 \\ \hline
           nomao &         118 &          2 &               23091 &           11374 \\ \hline
  bank-marketing &          16 &          2 &               30291 &           14920 \\ \hline
          isolet &         617 &         26 &                5223 &            2574 \\ \hline
     Bioresponse &        1776 &          2 &                2513 &            1238 \\ \hline
   mfeat-fourier &          76 &         10 &                1340 &             660 \\ \hline
   mfeat-factors &         216 &         10 &                1340 &             660 \\ \hline
       pendigits &          16 &         10 &                7364 &            3628 \\ \hline
       optdigits &          64 &         10 &                3765 &            1855 \\ \hline
         vehicle &          18 &          4 &                 566 &             280 \\ \hline
          cnae-9 &         856 &          9 &                 723 &             357 \\ \hline
        breast-w &           9 &          2 &                 468 &             231 \\ \hline
   balance-scale &           4 &          3 &                 418 &             207 \\ \hline
     SpeedDating &         120 &          2 &                5613 &            2765 \\ \hline
      eucalyptus &          19 &          5 &                 493 &             243 \\ \hline
           vowel &          12 &         11 &                 663 &             327 \\ \hline
 credit-approval &          15 &          2 &                 462 &             228 \\ \hline
          splice &          60 &          3 &                2137 &            1053 \\ \hline
             cmc &           9 &          3 &                 986 &             487 \\ \hline
\hline
\end{tabular}

       \vspace{1cm}
    \caption{OpenML datasets used for evaluation and comparison. We give number of features (\#f), the number of classes (\#c), as well as the size of the training and test sets for each benchmark.}
   \label{table:datasets}
\end{table}



\subsubsection{Base Models}
For each benchmark, we built a set of 5 base classifiers, Naive Bayes (NB), Support Vector Classification (SVC), Perceptron, $k$-Nearest Neighbors ($k$-NN), and Decision Tree Classifier (DTC). 
\subsubsection{Competing Algorithms}
We use the methods reviewed in the related works section for our comparison, with the exception of A Posteriori (APO) and KNORA-E (KE) which we found were significantly outperformed by their respective sister methods, APR and KU. Instead, we include a simple majority vote (MV) on the neighborhood instances. We use Python library DESLIB~\cite{deslib} which implements all competing methods. 

For each method, DESLIB uses 50\% of the training data to train the base classifiers, and the remaining training data to train the dynamic classification selection method. To make the comparison fair, we use only 50\% of the training data to create the clusterings in CSHC as well. Note that we limit CSHC in this way purely to level the playing field with the competing algorithms. In practice, one will want to use 100\% of the training data, labeled in one, or possibly multiple, cross-validation(s), to create the cost-sensitive clusters. 
%
\subsubsection{Hyper Parameters}
For Naive Bayes, we use class frequencies as priors. For the Support Vector Classification, we use an RBF kernel with $C=1$ and $\gamma = {1\over \#\mathrm{features}}$. For $k$-NN, we use a simple one-nearest neighbor classification ($k=1$). For the Decision Tree Classifier, finally, we use the Gini index as branching metric, impose no depth limit, and perform no pruning. 
%

CSHC has six hyper parameters. We generate 50 trees using the number generator from~\cite{random}. For each tree, we sample with repetition from the training set until we obtain a multi-set of samples which amounts to 80\% of the total training set of unique samples. To create each tree, we use two times the square root of all features, chosen uniformly at random. The last three hyper parameters determine when we stop the hierarchical refinement of clusters. First, we enforce that at least 2 samples remain within each cluster. Second, we limit the depth of the trees to be at most 15. And finally, we stop refining the clusters when the improvement by an additional split drops below 2\%.

For the LP-based weighting scheme, we set $\gamma=80$. The recourse threshold $\rho$ is set to $0.5$, which means that we only trust a classifier selection method outright when the support for the highest ranked class is at least twice that of the second most supported class. Note that all these parameters are set to the same values for all benchmarks we consider in the experiments. Naturally, these hyper-parameters could be tuned for each benchmark individually, for example by means of a cross validation. To demonstrate the effectiveness of the method proposed, we leave all CSHC parameters and the parameters for the modifications we introduced at the same default values for all benchmarks. 

For all other selection methods, we use the DESLIB library defaults for all hyper-parameters~\cite{deslib}.
\subsubsection{Hardware, Operating System, Compilers, and Libraries}
The experiments for CSHC and it's variants were performed on a 16-CPU cluster of  8-core, 2.60GHz  Intel(R) Xeon(R) CPU ES-2670 with a 20MB cache size. IBM ILOG CPLEX 12.6.3 was the solver used to solve the linear programs. The algorithms were coded in C++ using the GCC 4.8.5 compiler on a Red Hat 4.8.5-4 operating system. The numerical experiments with the competing algorithms were performed on a 6-core, 2.71Ghz Intel(R) Xeon(R) CPU E-2176M with 64MB RAM running a Windows operating system. The Python (3.7.7) library DESLIB v0.3~\cite{deslib} was used to implement the competing algorithms.


\begin{figure}[t!]
\vspace{-0.3cm}
\begin{center}
  \begin{subfigure}[b]{0.49\columnwidth}
    \includegraphics[width=1\columnwidth]{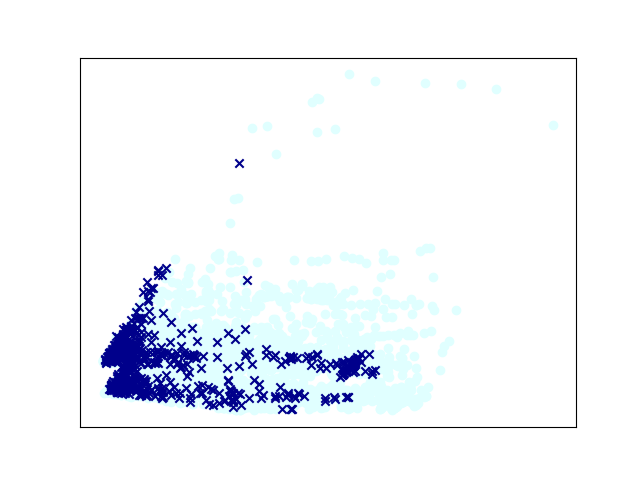}
    \caption{$k$-NN}
    \label{fig:2a}
  \end{subfigure}
  \hfill 
  \begin{subfigure}[b]{0.49\columnwidth}
    \includegraphics[width=1\columnwidth]{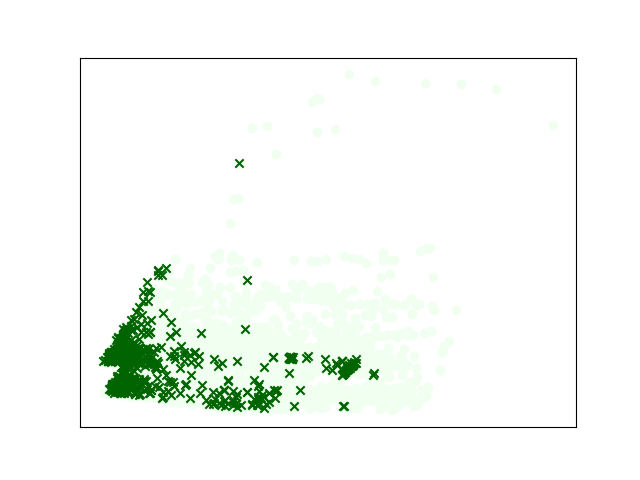}
    \caption{SVC}
    \label{fig:2b}
  \end{subfigure}
  \\
  \vspace{-0.1cm}
    \begin{subfigure}[b]{0.49\columnwidth}
    \includegraphics[width=1\columnwidth]{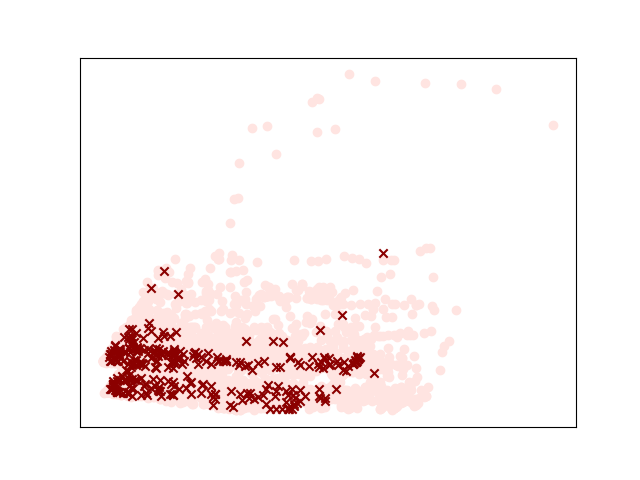}
    \caption{DT}
    \label{fig:2c}
  \end{subfigure}
  \hfill
    \begin{subfigure}[b]{0.49\columnwidth}
    \includegraphics[width=1\columnwidth]{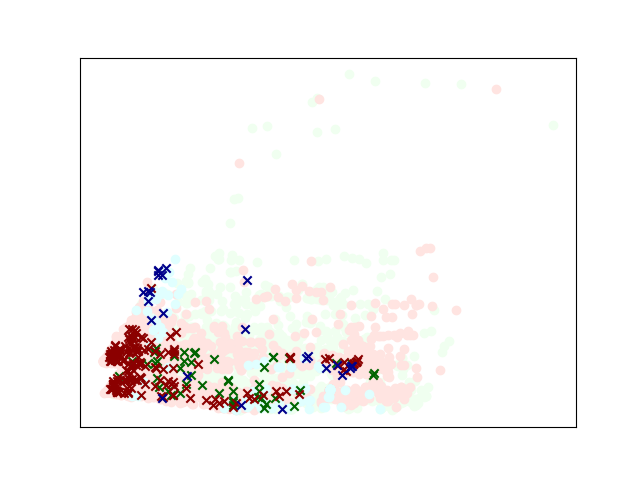}
    \caption{CSHC-LPR}
    \label{fig:2d}
  \end{subfigure}
  \caption[Predictions of different classifiers]{Predictions of different base classifiers as well as CSHC-LPR for the mozilla4 dataset. The lighter round markers and darker star markers indicate correct and incorrect predictions, respectively. In (d), CHSC-LPR selects a base classifier  dynamically  for each point. We use the same color coding as in (a)-(c) to show which classifier is selected. \label{fig:comparative}}
  \end{center}
\end{figure}

We illustrate the DCS concept in Figure~\ref{fig:comparative}. We plot the test cases of the mozilla4 benchmark set as projected into the two most significant principle components. We mark the error cases in bold for $k$-NN, SVC, and DT (we omitted NB and Perceptron to save space and because CSHC-LPR hardly ever chooses them on this benchmark). The CSHC-LPR tile shows in what region which classifier is selected.

\begin{table}[b!]
    \centering
    \begin{tabular}{|r|p{1cm}|p{1cm}|p{1cm}|p{1cm}||p{1cm}|}
\hline
{} &   \bf CSHC &  \bf   RR &   \bf  LP & \bf LPR & \bf Oracle \\ \hline \hline
balance-scale        &               90.3 &   90.3 &   90.3 &      90.3 &    93.2 \\
bank-marketing       &                89.7 &   89.5 &   89.4 &      89.5 &    96.5 \\
Bioresponse          &              74.4 &   75.4 &   75.8 &      75.8 &    93.7 \\
breast-w             &               97.4 &   97.0 &   97.4 &      97.0 &    98.7 \\
cmc                  &              52.2 &   51.1 &   49.9 &      51.3 &    85.2 \\
cnae-9               &              86.6 &   88.0 &   89.1 &      88.0 &    96.6 \\
credit-approval      &              86.0 &   89.0 &   88.6 &      88.6 &    95.6 \\
credit-g             &              75.2 &   74.5 &   74.5 &      75.2 &    94.8 \\
diabetes             &             74.8 &   76.4 &   74.8 &      75.2 &    89.8 \\
eeg-eye-state        &             93.0 &   90.1 &   93.1 &      93.1 &    99.9 \\
eucalyptus           &             58.8 &   62.6 &   60.5 &      61.7 &    86.4 \\
gina\_agnostic        &   89.3 &   88.6 &   89.3 &      89.3 &    98.0 \\
heart-h              &             81.6 &   80.6 &   80.6 &      80.6 &    89.8 \\
hill-valley          &         51.5 &   52.5 &   52.2 &      53.8 &    97.2 \\
ilpd                 &             70.5 &   71.0 &   72.5 &      72.5 &    99.0 \\
isolet               &             94.8 &   94.6 &   94.9 &      94.9 &    98.6 \\
kc1                  &             84.9 &   86.1 &   85.2 &      85.1 &    94.0 \\
kc2                  &         80.3 &   83.8 &   81.5 &      83.2 &    91.9 \\
kr-vs-kp             &            98.4 &   98.0 &   98.5 &      98.4 &   100.0 \\
letter               &        93.0 &   93.3 &   93.4 &      93.4 &    97.5 \\
mfeat-factors        &             95.8 &   96.4 &   96.5 &      96.2 &    98.8 \\
mfeat-fourier        &              79.5 &   80.6 &   80.9 &      80.3 &    93.9 \\
mozilla4             &             92.4 &   90.1 &   92.2 &      92.3 &    98.3 \\
musk                 &            100 &  100 &  100 &     100 &   100 \\
nomao                &             95.5 &   95.7 &   95.8 &      95.7 &    99.2 \\
optdigits            &             98.2 &   98.2 &   98.3 &      98.3 &    99.6 \\
ozone-level-8hr      &            92.6 &   93.1 &   93.1 &      92.7 &    98.6 \\
pc1                  &     95.6 &   95.6 &   95.4 &      95.6 &    97.8 \\
pc3                  &             90.1 &   89.5 &   89.0 &      89.5 &    95.3 \\
pendigits            &             99.3 &   99.2 &   99.3 &      99.3 &    99.7 \\
phoneme              &             84.4 &   86.0 &   85.2 &      85.4 &    97.8 \\
qsar-biodeg          &            84.2 &   84.2 &   85.7 &      85.1 &    96.6 \\
scene                &         95.5 &   95.5 &   84.5 &      96.0 &    99.5 \\
spambase             &     93.7 &   94.1 &   95.7 &      94.1 &    99.3 \\
SpeedDating          &             85.1 &   85.4 &   94.1 &      85.4 &    97.0 \\
splice               &             91.6 &   93.3 &   93.2 &      93.0 &    98.5 \\
tic-tac-toe          &              85.8 &   86.1 &   86.8 &      87.1 &    97.5 \\
vehicle              &             73.6 &   77.1 &   76.8 &      77.1 &    92.1 \\
vowel                &             83.5 &   82.6 &   85.6 &      84.4 &    94.5 \\
wdbc                 &             97.9 &   96.3 &   97.9 &      98.4 &    99.5 \\ 
\hline \hline
\# losses/*-LPR &              27 &   20 &   16 &       0 &     0.0 \\
\# wins/*-LPR   &           6 &   11 &   14 &       0 &    39 \\
MGI  [\%]                &    0.8 &  0.3 &  0.2 &     0 &   -10.9 \\
\hline
\end{tabular}

    \vspace{1cm}
    \caption{Comparing accuracy of CSHC variants with the best static classifier and the perfect Oracle.The last rows give the number of wins/losses over CSHC-LPR, and the geometric mean of the accuracy ratio of CSHC-LPR and the respective method, minus one (MGI).}
    \label{tab:results_CSHC}
\end{table}

\subsection{Highest Rank vs.\ Rank Regression}

We begin our experimentation by comparing vanilla CSHC with the rank regression scheme (CSHC-RR) we introduced. Recall that CSHC selects the classifier that has the highest average rank over all clusters the test sample falls into. The rank regression modification we introduced, on the other hand, uses these average ranks as weights for the support each classifier gives to their favorite class. 

The performances of the two methods are depicted in columns two and three in Table~\ref{tab:results_CSHC}. Using our five very basic classifiers, we build ensembles using vanilla CSHC and with our newly introduced rank regression. Please note that the objective of our experiments is not to create the best approach for each benchmark in absolute terms, but to compare the relative performance of different classifier selection methods. In fact, exactly because our base classifiers are crude and relatively weak, the classifier selection is more challenging, which is the setting we strive for when comparing different DCS methods. If all base classifiers returned mostly the correct labels anyway, it would be much harder to assess the effectiveness of DCS methods. 

We observe that, out of the 40 head-to-head comparisons, CSHC-RR wins 21 and loses 14, while on 5 benchmarks both methods perform equally well. This confirms our initial speculation that using the actual classifications of each classifier to select the top classifier gives an advantage. However, note that this additional performance comes at the cost of having to run all classifiers first. CSHC, on the other hand, selects one classifier based on the original features, and thus only requires one base classifier to run.


\subsection{Rank Regression vs.\ LP-based Weighting}

Next, we compare our new rank regression scheme with the more elaborate LP-based weighting which requires solving a linear program for each test sample, thereby making this method rather computationally expensive. We can infer from Table~\ref{tab:results_CSHC} that the LP-based method performs with higher accuracy on 21 benchmarks while performing worse on only 14. Moreover, the average accuracy is slightly higher as well.


\subsection{Using Recourse at Low Confidence}

In Table~\ref{tab:results_CSHC}, we also show the performance of the confidence-assessment-and-recourse process we introduced (CSHC-LPR). Recall from the earlier description that we run the rank regression first and, provided confidence is high, we move on directly with the classifier selected. Only when confidence is low, we employ the LP-based weighting method. If its confidence is high, we use the selected classifier, otherwise we consider the classifier with the highest average rank and eventually the majority class in the neighborhood to break the tie. 

The last four rows in the table compare every other method with this one. In the last row, we highlight the average rank over all benchmarks and datasets. In the last row, we show 
$MGI(X) \leftarrow GM\left( \frac{\mathrm{Accuracy_{CSHC-LPR}}}{\mathrm{Accuracy_X}}\right) - 1$,
in percent, where $X$ is the method compared with, and $GM(\cdot)$ is the geometric mean of a vector. Note that the $MGI(X)$ is the same as the ratio of the geometric mean of the accuracy of CSHC-LPR over the geometric mean of the accuracy of method $X$, minus 1. Therefore, a value greater zero indicates that method $X$ has lower accuracy than CSHC-LPR on average.  

We observe that using confidence assessments helps make the selection more robust. CSHC-LPR outperforms all other CSHC variants, both in terms of wins/losses over the 40 benchmarks, as well as in terms of the accuracy comparison as measured by MGI. At the same time, since most test samples can be handled by rank regression with high confidence, the method works considerably faster than CSHC-LP. The largest benchmark is bank-marketing with 16 features and close to 15,000 train and test samples each. Generating the 50 clusterings takes 9 sequential CPU seconds (KU 4s, MD 10s) which is manageable, especially since the training of different trees can be parallelized easily with linear speed-up.  
On hill-valley, which has 100 features and around 400 train and test samples, CSHC takes 0.47 seconds to train. (KU 0.04s, MD 0.29s), but therefore requires only 1.1 milliseconds to classify each sample (KU 75ms, MD 219ms).


The reason why CSHC-LPR is so efficient is because most test samples are handled by the fast rank regression technique, while KU and MD need to compute Euclidean distances to run $k$-NN in a high-dimensional space. On bank-marketing, e.g., only for 621, or 4.16\%, test samples confidence in the initial rank regression is too low and the LP-based weighting method is invoked. This means that, given the default threshold $\rho=0.5$, in over 95\% of the test cases the support of the top class is at least twice as high as the class with second largest support, which in turn implies that the support for the top class is at least 66\%. This illustrates how effective CSHC is at partitioning the feature space in such a way that the choice of classifier is clear.

\begin{table}[b!]
    \centering
    \begin{tabular}{|p{1.5cm}|p{1cm}|p{1cm}|p{1cm}|p{01cm}|p{1cm}|p{1cm}|p{1cm}|p{1cm}|
}
\hline
{} &   APR &   MCB &    OLA &    MV &     MD &   CSHC &    KU &  *LPR \\ \hline \hline

balance-*        &   87.0 &   86.0 &   88.9 &   89.9 &   87.9 & \bf  90.3 &   89.4 &  \bf    90.3 \\ \hline
bank-ma*       &   88.6 &   88.8 &   89.0 &   89.2 &   89.2 &  \bf 89.7 &   89.5 &      89.5 \\ \hline
Bioresp*        &   72.9 &   73.4 &   72.3 &   75.1 &   72.2 &   74.4 &   75.1 &    \bf  75.8 \\ \hline
breast-w*             &   95.2 &   95.2 &   95.7 &   96.5 &   95.7 &  \bf 97.4 &   96.1 &      97.0 \\ \hline
cmc                  &   47.2 &   49.1 &  \bf 52.8 &   49.3 &   48.3 &   52.2 &   52.4 &      51.3 \\ \hline
cnae-9               &   82.1 &   84.6 &   85.7 &   88.5 &   88.2 &   86.6 &   \bf 89.1 &      88.0 \\ \hline
credit-ap*      &   85.5 &   85.1 &   87.3 &   88.6 &   87.3 &   86.0 &  \bf 89.5 &      88.6 \\ \hline
credit-g             &   73.3 &   73.9 &   76.7 &   74.8 &   74.2 &   75.2 &  \bf 77.0 &      75.2 \\ \hline
diabetes             &   72.4 &   73.2 &   73.6 & \bf  76.8 &   75.6 &   74.8 &   76.8 &      75.2 \\ \hline
eeg-eye-st*        &   91.8 &   92.7 &   92.1 &   87.5 &   92.3 &   93.0 &   92.5 &     \bf 93.1 \\ \hline
eucalypt           &   56.8 &   56.8 &   61.3 &   58.8 &   61.3 &   58.8 & \bf  63.4 &      61.7 \\ \hline
gina\_agn*        &   85.4 &   83.8 &   86.7 &   88.2 &   88.6 &  \bf 89.3 &   88.1 &    \bf  89.3 \\ \hline
heart-h              &  \bf 82.7 &   81.6 &  \bf 82.7 &   81.6 &   81.6 &   81.6 &   80.6 &      80.6 \\ \hline
hill-val*          &   50.0 &   49.8 &   47.8 &   53.5 &   51.2 &   51.5 &   49.8 &     \bf 53.8 \\ \hline
ilpd              &   69.4 &   66.8 &   68.4 &   \bf 72.5 &   63.7 &   70.5 &   71.5 &     \bf 72.5 \\ \hline
isolet*               &   89.0 &   91.1 &   93.1 &   94.1 &   94.6 &   94.8 &   94.5 &     \bf 94.9 \\ \hline
kc1                  &   86.5 &   86.1 &   85.5 &   86.2 &   85.8 &   84.9 &  \bf 86.9 &      85.1 \\ \hline
kc2                  &   76.9 &   78.6 &   79.8 &  \bf 85.5 &   82.1 &   80.3 &   82.7 &      83.2 \\ \hline
kr-vs-kp            &   97.9 &   97.6 &   97.7 &   97.6 &  \bf 98.4 &  \bf 98.4 &   98.2 &    \bf  98.4 \\ \hline
letter*               &   91.8 &   91.8 &   92.2 &   90.9 &   92.5 &   93.0 &   93.4 &     \bf 93.4 \\ \hline
mfeat-fac*        &   94.7 &   94.5 &   95.5 &   95.6 &  \bf 96.5 &   95.8 &   96.4 &      96.2 \\ \hline
mfeat-fou*        &   77.7 &   78.5 &   78.5 &   81.2 &  \bf 81.4 &   79.5 &   80.2 &      80.3 \\ \hline
mozilla4            &   90.9 &   90.9 &   91.4 &   88.5 &   91.8 &   \bf 92.4 &   88.8 &      92.3 \\ \hline
musk                 &   99.4 &   99.5 &   99.9 &   99.7 & \bf 100 & \bf 100 &   99.8 &   \bf  100 \\ \hline
nomao                &   95.4 &   95.3 &   95.5 &   95.5 &   95.8 &   95.5 &  \bf 95.8 &      95.7 \\ \hline
optdigits            &   96.5 &   96.3 &   96.7 &   97.9 &  \bf 98.3 &   98.2 &   97.9 &      98.3 \\ \hline
ozone-*      &   92.2 &   93.3 &  \bf 93.5 &   93.1 &   93.4 &   92.6 &   93.2 &      92.7 \\ \hline
pc1                  &   93.7 &   94.5 &   94.3 &   95.1 &   94.3 &   \bf 95.6 &   95.1 &    \bf  95.6 \\ \hline
pc3                  &   88.2 &   87.6 &   88.6 &   89.1 &   89.0 &  \bf  90.1 & \bf  90.1 &      89.5 \\ \hline
pendig*            &   98.5 &   98.4 &   97.8 &   98.6 &   99.2 &   \bf99.3 &   99.0 &     \bf 99.3 \\ \hline
phoneme              &   86.5 &   86.5 &   \bf 87.0 &   84.8 &   86.7 &   84.4 &   86.2 &      85.4 \\ \hline
qsar-bio*          &   82.5 &   83.4 &   83.4 &   83.4 &  \bf 85.7 &   84.2 &   84.8 &      85.1 \\ \hline
scene                &   93.2 &   94.0 &   95.7 &   95.2 & \bf   96.6 &   95.5 &   95.3 &      96.0 \\ \hline
spambase             &   92.2 &   93.0 &   93.3 &  \bf 94.1 &   93.7 &   93.7 &   93.7 &     \bf 94.1 \\ \hline
SpeedDat*          &   83.6 &   83.8 &   84.5 &   85.0 &   83.8 &   85.1 &   85.2 &     \bf 85.4 \\ \hline
splice               &   89.9 &   90.3 &   91.4 &   92.7 &   93.3 &   91.6 &   \bf 93.4 &      93.0 \\ \hline
tic-tac*          &   84.9 &   85.2 &   86.1 &   81.1 &   83.9 &   85.8 &   81.4 &    \bf  87.1 \\ \hline
vehicle*              &   71.4 &   72.1 &   75.0 &   75.0 &   74.6 &   73.6 &   76.8 &  \bf    77.1 \\ \hline
vowel                &   87.5 &   84.4 &   86.2 &   80.4 &  \bf  90.2 &   83.5 &   82.9 &      84.4 \\ \hline
wdbc*                 &   95.2 &   95.2 &   97.3 &   96.3 &   97.9 &   97.9 &   97.9 &    \bf  98.4 \\ \hline
\hline \hline
losses/LPR  &   36 &   35 &   33 &   33 &   24 &   27 &   26 &       0 \\ \hline
wins/LPR    &    4 &    4 &    7 &    7 &   14 &    6 &   13 &       0 \\ \hline
MGI [\%]                 &  2.8 &  2.6 &  1.3 &  1.1 &  1.0 &  0.8 & 0.4 &     0 \\ \hline
rank                 &   2.3  &    2.6 &    4.1 &    4.6  &    5.2 &    5.3 &    5.6  &       6.4 \\ \hline
\end{tabular}

    \vspace{1cm}
    \caption{Comparing the accuracy of CSHC and CSHC-LPR with other state-of-art methods. The last rows give the number of wins/losses over CSHC-LPR, the MGI, and the average ranks over all methods and benchmarks (the higher the rank, the higher the relative performance). The names of the competing methods are given in the related works section.}
    
    \label{tab:results}
\end{table}

Most of the overrides result in no change. In 318 cases, both *-RR and *-LPR choose a classifier that classifies the input correctly, in 263 cases both select a classifier that errs on the respective input. In 15 cases, the override worsens the outcome: *-RR would have chosen a good classifier, but *-LPR chooses one that is favors the wrong class. However, in 25 cases the initial rank regression would have chosen a bad classifier, but the recourse corrected that mistake and chose a classifier that labels the input correctly. In total, the rank regression method errs on 1,573 out of 14,919 test samples, 288, or 18.3\%, of these mistakes happen on cases for which the recourse is invoked. That implies that the error rate is over 4.4 times higher on recourse cases than on cases where we trust our primary rank regression method. This shows that our support-ratio recourse indicator is quite effective at identifying problematic cases. However, there is clearly room to improve the actual recourse which only gives us a net gain of 10 test samples. 

\subsection{Comparison With the State of the Art}

In Table~\ref{tab:results}, we compare the original CSHC and CSHC-LPR with prominent DCS methods from the literature, whereby KNORA-U and META-DES are widely regarded as the current state of the art. As before, in the last rows we show wins/losses and MGI when compared with CSHC-LPR. In the last row we also give the average rank over all 40 benchmarks when ranking all eight methods for each benchmark individually. This data confirms that KNORA-U and META-DES are outstanding dynamic classifier selectors. Surprisingly, we find that simple majority voting is a close runner-up to these sophisticated selection methods. 

\medskip

Regarding CSHC-LPR, we see that it compares very favorably with all other methods. Even KNORA-U, the strongest competitor in terms of average accuracy, is outperformed on 26 out of 40 benchmarks, and tied on one (namely heart-h - note that we round accuracy in the table, hence sometimes two methods may appear to have the same performance when they do not. On the letter benchmark, for example, CSHC-LPR actually outperforms KU). Running a paired student t-test based on this win/loss data results in a p-value of 3.56\% for the Null-hypothesis that both methods performed equally well, which allows us to refute this assumption with statistical significance at the commonly applied significance level of 5\%.

Furthermore, we can observe from this table that the original, unmodified CSHC method is almost as good as the best DCS methods to date. Compared with KNORA-U, it performs better on 18 benchmarks and worse on 20, with 2 benchmarks tied. In practice, this makes CSHC a very attractive choice, since it does not require running all base classifiers, but only the one it selects. This means that an ensemble of classifiers can be used effectively for boosting accuracy without having to pay a significantly higher computational cost which may be attractive for keeping the energy and $\mathrm{CO}_2$ burden low for mass applications.


\section{Conclusion}
We studied the use of cost-sensitive hierarchical clustering (CSHC) for the purpose of dynamic classifier selection (DCS) in ensemble learning.  We introduced two modifications of CSHC, one based on a rank-based weighting of classifications, the other using an input-specific linear programming formulation to compute a convex combination of classifications. We also introduced a confidence assessment and recourse process to decide which selection method to trust. Experimental results on 40 established machine learning benchmarks with fixed hyper-parameters showed that the modified CSHC works robustly and favorably when compared to various other DCS methods from the literature. 

Our project dictated that we must choose one out of a set of classifiers to enable the inheritance of explanations from base classifiers. It is worth noting, though, that CSHC can also be adjusted to aggregate the scores of individual classes by multiple classifiers so that it also presents an alternative to traditional stacking techniques as well. This, as well as the use of augmenting the feature set for cost-sensitive hierarchical clustering, are part of our future work.

\newpage

\bibliography{ref}

\begin{thebibliography}{10}

\bibitem{valiant84}
Leslie~G. Valiant.
\newblock A theory of the learnable.
\newblock {\em Commun. {ACM}}, 27(11):1134--1142, 1984.

\bibitem{freund_short_nodate}
Yoav Freund and Robert~E Schapire.
\newblock A decision-theoretic generalization of on-line learning and an
  application to boosting.
\newblock {\em Journal of Computer and System Sciences}, 55(1):119 -- 139,
  1997.

\bibitem{wolpert_stacked_1992}
David~H. Wolpert.
\newblock Stacked {Generalization}.
\newblock {\em Neural Networks}, pages 241--259, 1992.

\bibitem{leyton-brown_portfolio_2003}
Kevin Leyton-Brown, Eugene Nudelman, Galen Andrew, Jim McFadden, and Yoav
  Shoham.
\newblock A portfolio approach to algorithm select.
\newblock In {\em Proceedings of the 18th international joint conference on
  {Artificial} intelligence}, {IJCAI}'03, pages 1542--1543, San Francisco, CA,
  USA, August 2003. Morgan Kaufmann Publishers Inc.

\bibitem{sat_competition}
Balint, Belov, Heule, and Järvisalo.
\newblock Proceedings of sat competition 2013 : Solver and benchmark
  descriptions.
\newblock University of Helsinki , Helsinki, Finland, 2013.

\bibitem{malitskyAlg2013}
Yuri Malitsky, Ashish Sabharwal, Horst Samulowitz, and Meinolf Sellmann.
\newblock Algorithm portfolios based on cost-sensitive hierarchical clustering.
\newblock In Francesca Rossi, editor, {\em {IJCAI} 2013, Proceedings of the
  23rd International Joint Conference on Artificial Intelligence, Beijing,
  China, August 3-9, 2013}, pages 608--614, 2013.

\bibitem{woods_combination_1997}
K.~Woods, W.P. Kegelmeyer, and K.~Bowyer.
\newblock Combination of multiple classifiers using local accuracy estimates.
\newblock {\em IEEE Transactions on Pattern Analysis and Machine Intelligence},
  19(4), April 1997.

\bibitem{giacinto_methods_1999}
G.~Giacinto and F.~Roli.
\newblock Methods for dynamic classifier selection.
\newblock In {\em Proceedings 10th {International} {Conference} on {Image}
  {Analysis} and {Processing}}, September 1999.

\bibitem{giacinto_dynamic_2001}
Giorgio Giacinto and Fabio Roli.
\newblock Dynamic classifier selection based on multiple classifier behaviour.
\newblock {\em Pattern Recognition}, 34, September 2001.

\bibitem{ko_dynamic_2008}
Albert H.~R. Ko, Robert Sabourin, and Jr. Britto, Alceu~Souza.
\newblock From dynamic classifier selection to dynamic ensemble selection.
\newblock {\em Pattern Recognition}, 41(5), May 2008.

\bibitem{cruz_meta_2015}
Rafael M.~O. Cruz, Robert Sabourin, George D.~C. Cavalcanti, and Tsang Ing~Ren.
\newblock {META}-{DES}: {A} dynamic ensemble selection framework using
  meta-learning.
\newblock {\em Pattern Recognition}, 48(5), may 2015.

\bibitem{cruz_dynamic_2018}
Rafael M.~O. Cruz, Robert Sabourin, and George D.~C. Cavalcanti.
\newblock Dynamic classifier selection: {Recent} advances and perspectives.
\newblock {\em Information Fusion}, 41, May 2018.

\bibitem{AnsoteguiST18}
Carlos Ans{\'{o}}tegui, Meinolf Sellmann, and Kevin Tierney.
\newblock Self-configuring cost-sensitive hierarchical clustering with
  recourse.
\newblock In {\em Principles and Practice of Constraint Programming - 24th
  International Conference, {CP} 2018, Lille, France, August 27-31, 2018,
  Proceedings}, Lecture Notes in Computer Science, pages 524--534. Springer,
  2018.

\bibitem{feurer_openml-python_2019}
Matthias Feurer, Jan~N. van Rijn, Arlind Kadra, Pieter Gijsbers, Neeratyoy
  Mallik, Sahithya Ravi, Andreas Müller, Joaquin Vanschoren, and Frank Hutter.
\newblock {OpenML}-{Python}: an extensible {Python} {API} for {OpenML}.
\newblock {\em arXiv:1911.02490}, November 2019.

\bibitem{vanschoren_openml_2014}
Joaquin Vanschoren, Jan~N. van Rijn, Bernd Bischl, and Luis Torgo.
\newblock {OpenML}: networked science in machine learning.
\newblock {\em ACM SIGKDD Explorations Newsletter}, 15, June 2014.

\bibitem{deslib}
Rafael M.~O. Cruz, Luiz~G. Hafemann, Robert Sabourin, and George D.~C.
  Cavalcanti.
\newblock Deslib: A dynamic ensemble selection library in python.
\newblock {\em Journal of Machine Learning Research}, 21(8):1--5, 2020.

\bibitem{random}
Agner Fog.
\newblock Pseudo random number generators uniform and non-uniform
  distributions.
\newblock \url{https://www.agner.org/random/}, 2019.

\end{thebibliography}

\bigskip




\end{document}